%% file: main.tex
\documentclass[lettersize,journal]{IEEEtran}
\usepackage{amsmath,amsfonts}
\usepackage{algorithmic}
\usepackage{algorithm}
\usepackage{array}
\usepackage{textcomp}
\usepackage{stfloats}
\usepackage{url}
\usepackage{verbatim}
\usepackage{graphicx}
\usepackage{cite}
\input{macros}

\begin{document}

\title{
An Improved Template for Approximate Computing
}

\author{
Morteza Rezaalipour,
Francesco Costa,
Marco Biasion,\\
Rodrigo Otoni,
George A. Constantinides,
and Laura Pozzi
\thanks{Morteza Rezaalipour, Francesco Costa, Marco Biasion, Rodrigo Otoni, and Laura Pozzi are all with the Faculty of Informatics, Universit{\`a} della Svizzera italiana, Lugano, Switzerland (e-mail: morteza.rezaalipour@usi.ch; francesco.costa@usi.ch; marco.biasion@usi.ch; otonir@usi.ch; laura.pozzi@usi.ch).}
\thanks{George A. Constantinides is with the Faculty of Engineering, Imperial College London, London, UK (e-mail: g.constantinides@imperial.ac.uk).}
\thanks{This work was partially supported by the Swiss National Science Foundation via grant 200020\_188613.}
}

\markboth{}%
{Rezaalipour \MakeLowercase{\textit{et al.}}: Template Proxies for Approximate Logic Synthesis}

\maketitle


\input{abstract}

\input{section_introduction}
\input{section_templates}
\input{section_proxy}
\input{section_evaluation}
\input{section_conclusion}


\bibliographystyle{IEEEtran}
\bibliography{references}

\end{document}

%% file: macros.tex
\usepackage{acronym}
\usepackage{subcaption}
\usepackage{multirow}
\usepackage{booktabs}
\usepackage{amssymb} 
\usepackage[normalem]{ulem} 
\usepackage[colorlinks=true, allcolors=blue]{hyperref}

\newacro{sat}[SAT]{Boolean satisfiability}
\newacro{smt}[SMT]{satisfiability modulo theories}
\newacro{qbf}[QBF]{quantified Boolean formula}
\newacro{cdcl}[CDCL]{conflict-driven clause learning}
\newacro{adt}[ADT]{abstract data type}
\newacro{fol}[FOL]{first-order logic}
\newacro{lia}[LIA]{linear integer arithmetics}
\newacro{bv}[BV]{bit-vectors}
\newacro{et}[ET]{error threshold}
\newacro{als}[ALS]{approximate logic synthesis}
\newacro{lac}[LAC]{local approximation candidate}
\newacro{lut}[LUT]{lookup table}
\newacro{lpp}[\textsc{LPP}]{literals per product}
\newacro{ppo}[\textsc{PPO}]{products per output}
\newacro{pit}[\textsc{PIT}]{products in total}
\newacro{its}[\textsc{ITS}]{inputs to sums}
\newacro{nn}[NN]{neural network}

\newcommand{\tool}[1]{\textsc{#1}}

\newcommand{\xpat}{\tool{XPAT}}

\newcommand{\muscat}{\tool{MUSCAT}}

\newcommand{\mecals}{\tool{MECALS}}

\newcommand{\zthree}{\tool{Z3}}
\newcommand{\yosys}{\tool{Yosys}}
\newcommand{\shared}{\tool{SHARED}}

\newcommand{\blockBegin}[1][c]{\begin{array}{#1}}
\newcommand{\blockEnd}{\end{array}}
\newcommand{\blockEndDot}{\end{array} \!\!\! \right.}
\newcommand{\parBegin}[1][c]{\left ( \!\!\! \begin{array}{#1}}
\newcommand{\parBeginL}[1][l]{\left ( \!\!\! \begin{array}{#1}}
\newcommand{\parEnd}{\end{array} \!\!\! \right)}
\newcommand{\curlyBegin}[1][c]{\left \{ \!\!\! \begin{array}{#1}}
\newcommand{\curlyEnd}{\end{array} \!\!\! \right\}}

%% file: abstract.tex
\begin{abstract}
Deploying neural networks on edge devices entails a careful balance between the energy required for inference and the accuracy of the resulting classification. One technique for navigating this tradeoff is approximate computing: the process of reducing energy consumption by slightly reducing the accuracy of arithmetic operators. In this context, we propose a methodology to reduce the area of the small arithmetic operators used in neural networks -- i.e., adders and multipliers -- via a small loss in accuracy, and show that we improve area savings for the same accuracy loss w.r.t. the state of the art. To achieve our goal, we improve on a boolean rewriting technique recently proposed, called XPAT, where the use of a parametrisable template to rewrite circuits has  proved to be highly beneficial. In particular, XPAT was able to produce smaller circuits than comparable approaches while utilising a naive sum of products template structure. In this work, we show that template parameters can act as proxies for chosen metrics and we propose a novel template based on parametrisable product sharing that acts as a close proxy to synthesised area. We demonstrate experimentally that our methodology converges better to low-area solutions and that it can find better approximations than both the original XPAT and two other state-of-the-art approaches.
\end{abstract}


%% file: section_introduction.tex
\section{Introduction} \label{sec:introduction}


\IEEEPARstart{T}{he} deployment of \acp{nn} on edge devices entails a careful balance between the energy required for inference and the accuracy of the resulting classification. Small-bitwidth operators are often employed in \acp{nn} in order to limit required energy, e.g., 4-bit multipliers are used in~\cite{VenkataramaniJun21}, and the capability to \emph{further} reduce energy consumption by approximating such operators while minimally reducing their accuracy is critical. Approximate computing is a design paradigm which helps navigating this tradeoff, dictating that inexact hardware should be used whenever a loss in accuracy can be tolerated, in order to achieve improvements in metrics such as circuit area and energy consumption. In the context of \acp{nn},  efficient approximation of the employed arithmetic circuits, notably multipliers, is key to achieving good-enough inference results while keeping energy consumption at bay~\cite{ChenLL:2022}.

The process of deriving an approximate circuit from an exact circuit and a tolerated maximum \ac{et} is termed \ac{als}. Extensive work has been done in the last decade towards improving \ac{als}~\cite{
Scarabottolo20,
MrazekJun19,
ScarabottoloAP:2018,
HashemiJun18,
HashemiJan22,
WitschenMar22,
MengMar23},
but effectively exploiting the full flexibility allowed by the given \ac{et} is still a challenge. The recently proposed Boolean rewriting algorithm \xpat{}~\cite{Rezaalipour23a} was shown to be able to find valid high-quality optimisation candidates for a variety of arithmetic circuits. \xpat{} makes use of a \emph{parametrisable template} to represent the design space of potential optimisations and employs a \ac{smt} solver to traverse this large design space. The template proposed consisted of a sum of products, with the role of the solver being to identify, for every circuit output, which products of which input literals must be included in the synthesised circuit.

The choice of template has a significant effect on optimisation quality, since its parameters essentially guide \xpat's design space traversal. Thus, a good template structure is crucial. In particular, a template whose parameters act as a close \emph{proxy} for the metric of interest is likely to yield the best results. In light of this, we propose a novel parametrisable template for \ac{als} which is capable of sharing products' outputs among many sums as a means of \emph{avoiding low-quality optimisations during the search}. Our template's parameters regulate the inclusion and sharing of products and allow for fine-grained control of solver-based synthesis.

We implemented our methodology and compared it against the original version of \xpat{} and the state-of-the-art methods \muscat{}~\cite{WitschenMar22} and \mecals{}~\cite{MengMar23}. The results indicate that our shared template acts as a better proxy for circuit area than the original template, and that it yields better adder and multiplier approximations than all compared approaches.

\IEEEpubidadjcol 

To summarise, our contributions are the following:

\begin{itemize}
    \item[(1)] A novel parametrisable sum of products template for \ac{als} based on the sharing of products outputs.
    \item[(2)] A study showing that templates' parameters can act as proxies for metrics of interest such as circuit area.
    \item[(3)] An open-source implementation of the proposed methodology instrumented with our novel template.
    \item[(4)] An evaluation showcasing that our novel template acts as a good proxy for circuit area and yields better approximations than state-of-the-art approaches.
\end{itemize}

%% file: section_templates.tex
\section{Templates} \label{sec:templates}


We introduce the background needed by providing a summary of \xpat{}, followed by details of its original template. Then, we present the shared template proposed in this paper.


\subsection{XPAT Algorithm} \label{sec:xpat}

\begin{figure}[t]
	\centering
    \includegraphics[width=0.45\textwidth]{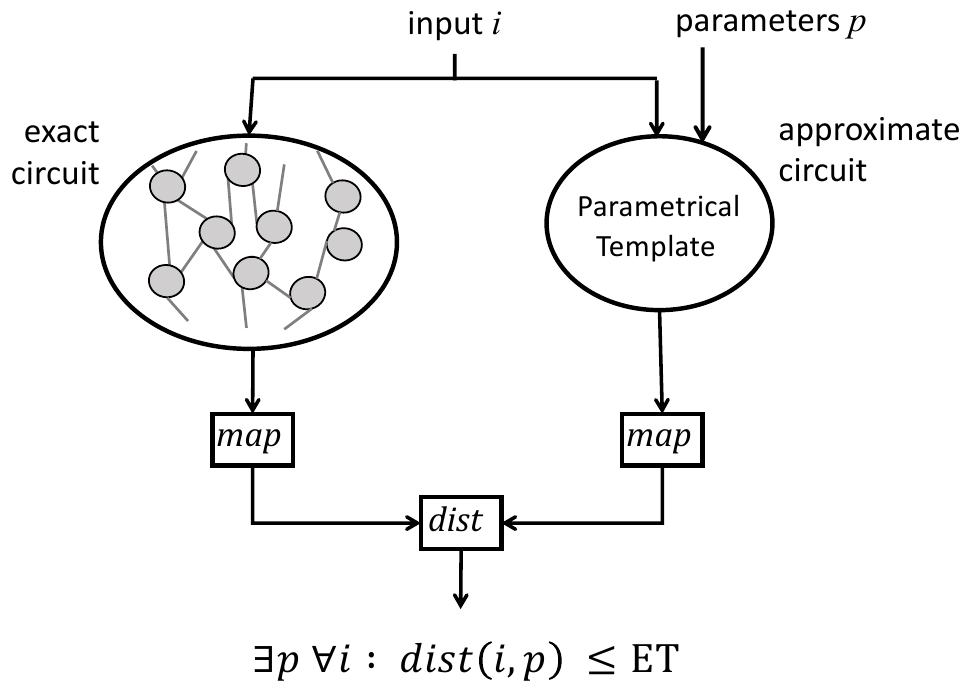}
    \caption{Miter for \xpat{}~\cite{Rezaalipour23a}.}
    \label{fig:xpat_miter}
\end{figure}

The algorithm frames the design space exploration in terms of \ac{smt} formulas, whose solving leads to the setting of the template's parameters in such a way that the approximate circuit respects the given \ac{et}. Concretely, an error miter is used, which consists of the exact circuit, the parametrisable template for approximation, two functions used for error estimation, and a query, as illustrated in Fig.~\ref{fig:xpat_miter}. Function $map$ does the mapping from the output of each circuit to an interpretation of the Boolean values and function $dist$ does the measurement of the distance between the mapped outputs of the exact and approximate circuits. The query asks the solver if the template parameters can be instantiated in a way that, for all possible input combinations, the distance between the exact and approximate outputs does not exceed the \ac{et}. A satisfiable result will be accompanied by an assignment to $p$ that ensures that the approximate circuit always respects the \ac{et}.

\subsection{Nonshared Template} \label{sec:nonshared_template}

\begin{figure}[t]
	\centering
    \includegraphics[width=0.45\textwidth]{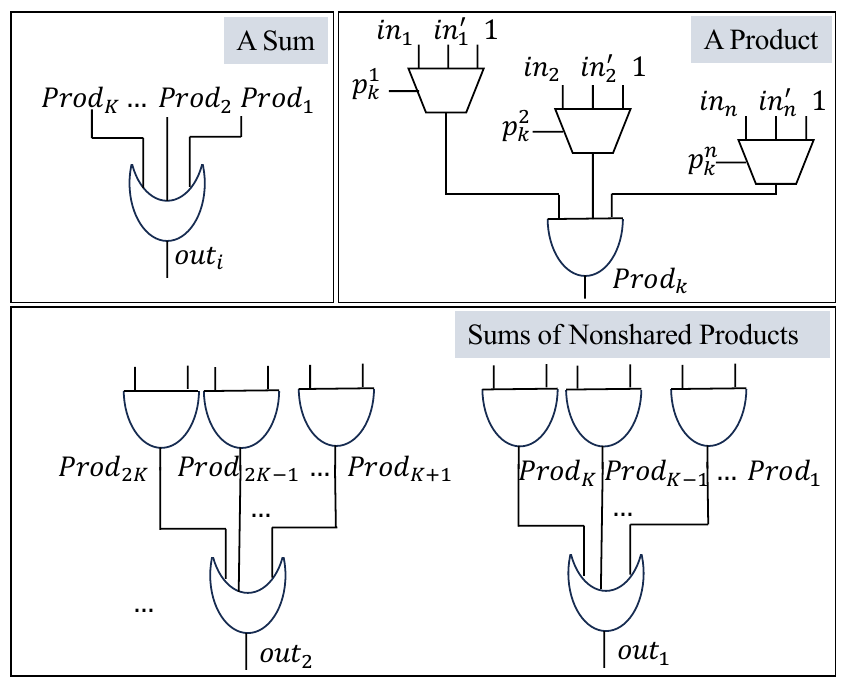}
    \caption{Template with nonshared sums of products.}
    \label{fig:template-nonshared}
\end{figure}

The template originally used by \xpat{} consists of \emph{nonshared} sums of products, with each sum dictating the value of one output and the parameters deciding which products of which inputs should be included in each sum, as illustrated in Fig.~\ref{fig:template-nonshared}. Formally, consider a given circuit with $n$ inputs, $in_1,...,in_n$, and $m$ outputs, $out_1,...,out_m$. The behaviour of each output $out_i$ is formalised as follows:
\begin{align}
    out_i = \bigvee_{k \in \{1...K\}} Prod_k
\end{align}
where $K$ is the number of products included in each sum, $Prod_k$ is a product of selected inputs, and $1 \leq i \leq m$. Each product $Prod_k$ has $n$ inputs, originating from $n$ multiplexers. Each multiplexer decides, based on parameter $p_k^j$, if input $in_j$ is used as is, is negated, or is replaced by the constant $1$, with $1 \leq j \leq n$; the usage of constant $1$ indicates that the respective input is ignored by the product\footnote{Note that the multiplexers are not a part of the generated circuit, since the selected lines are chosen at design time.}.

The strength of this template is its expressiveness, which guarantees \xpat{} to be able to represent any Boolean function. However, its limitation consists in the fact that \emph{every sum is conceived as an isolated element} having its own set of $K$ products, and this has tangible disadvantages. In fact, the template does not capture the concept of sharing products among outputs; as a result, product-sharing circuits may be overlooked in favour of equivalent approximations (from the \ac{smt} solver's perspective) in which all sums rely on disjoint sets of products.

\begin{figure}[t]
	\centering
    \includegraphics[width=0.45\textwidth]{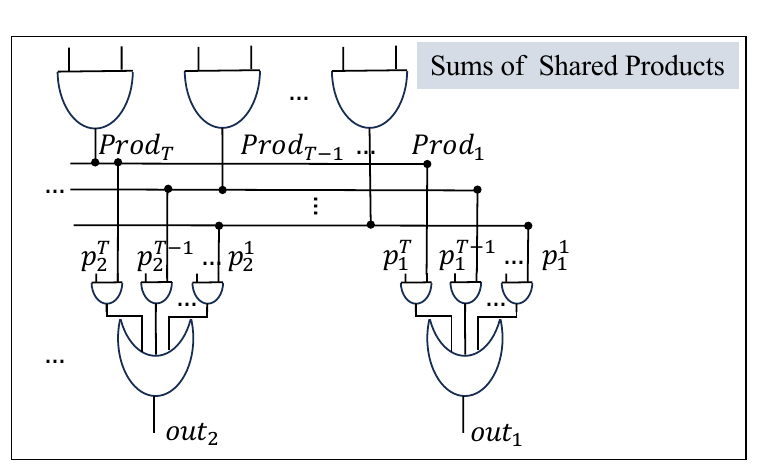}
    \caption{Template with shared sums of products.}
    \label{fig:template-shared}
\end{figure}


\subsection{Shared Template} \label{sec:shared_template}

The template we propose in this paper consists of sums of products in which the same product can be \emph{shared} among many sums, with new parameters deciding how the products should be shared, as illustrated in Fig.~\ref{fig:template-shared}. The behaviour of each output $out_i$ is formalised as follows:
\begin{align}
    out_i = \bigvee_{t \in \{1...T\}} Prod_t \lor \top
\end{align}
where $T$ is the total number of products in the template. Each output sum $out_i$ is associated with $T$ parameters $p_i^t$, which decide if product $t$ will be considered in $out_i$ or not.

Our template maintains the level of expressiveness of the nonshared template while allowing all $T$ products to be shared among the sums. This makes product sharing a factor in the exploration, sidestepping a large number of low-quality (mirrored) approximations from the design space.

%% file: section_proxy.tex
\section{Approximation Proxies} \label{sec:proxy}


Simply asking the \ac{smt} solver for a satisfying assignment might lead \xpat{} to a low-quality solution, since the solver is not able to differentiate between assignments that will lead to larger or smaller synthesised areas. To address this, the design space is systematically explored, with the aim of restricting the number of gates allowed in the approximate circuit, i.e., the search starts with a strong restriction, which is progressively weakened until an assignment is found. The shape of the restriction used, i.e., the constraints on the number and type of gates allowed, guides the exploration.

Since different templates have different parameters, they allow for different ways to guide the design space exploration. In particular, certain circuit features can act as a proxy for synthesised area. With the nonshared template, \xpat{}  restricted the number of \ac{lpp} and \ac{ppo} in its search for an approximate circuit. These constraints limit the number and size of the products that independently feed into any given sum. With our shared template, on the other hand, we can restrict the number of \ac{pit} and \ac{its}. These alternative constraints limit the total number of products in the whole circuit and the degree in which their outputs can be shared among the sums. As shown in the next section, \ac{pit} and \ac{its} act as a close proxy for synthesised area.

\begin{figure*}[t]
	\centering
	\begin{subfigure}[b]{0.44\textwidth}
		\centering
		\includegraphics[width=\textwidth]{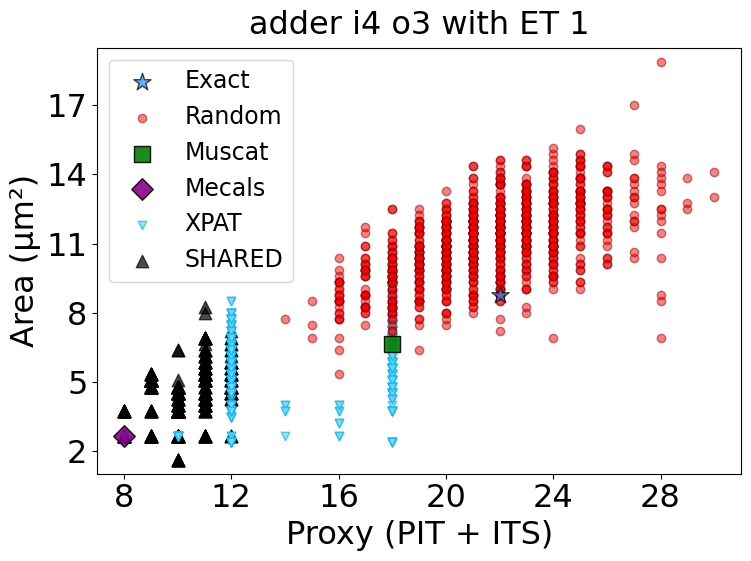}
	\end{subfigure}
    \hfill
	\begin{subfigure}[b]{0.44\textwidth}
		\centering
		\includegraphics[width=\textwidth]{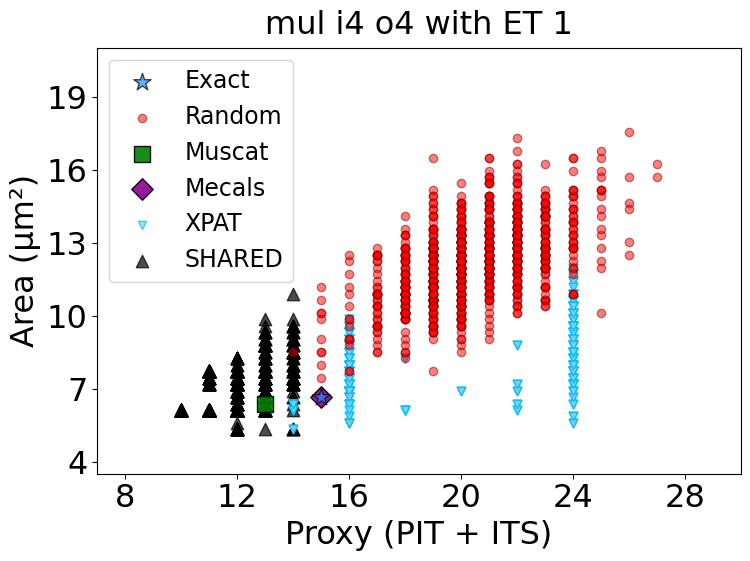}
	\end{subfigure}
    \hfill
    \begin{subfigure}[b]{0.44\textwidth}
		\centering
		\includegraphics[width=\textwidth]{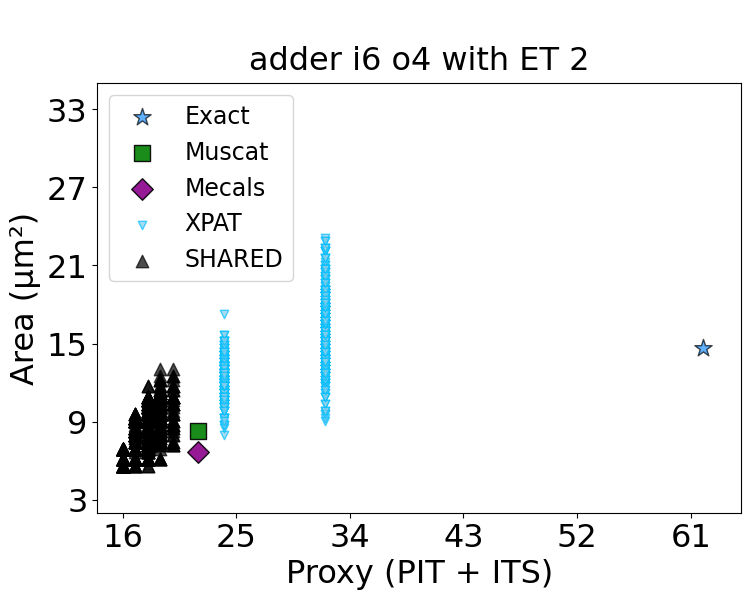}
	\end{subfigure}
    \hfill
	\begin{subfigure}[b]{0.44\textwidth}
		\centering
		\includegraphics[width=\textwidth]{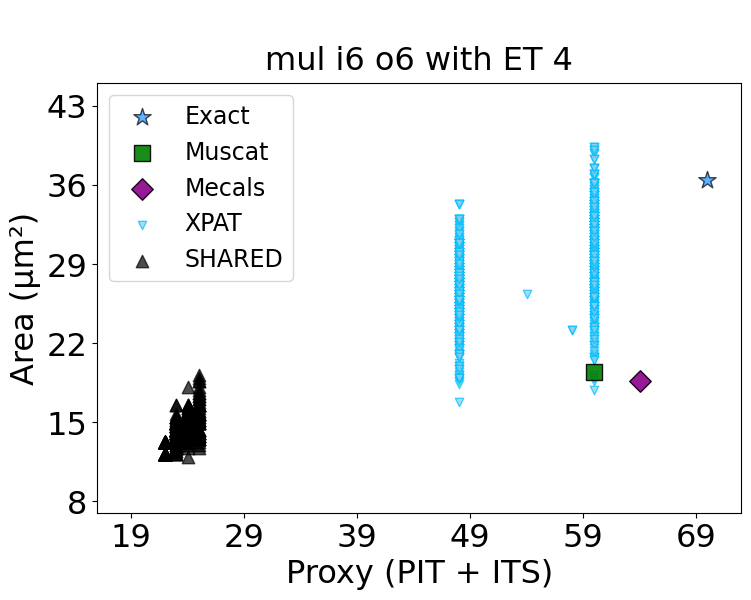}
	\end{subfigure}
    \hfill
	\caption{Area plots for approximate adder and multiplier circuits, showing area in relation to proxy values, for a fixed \ac{et}.}
	\label{fig:area_plots_clouds}
\end{figure*}

%% file: section_evaluation.tex
\section{Evaluation} \label{sec:evaluation}


\begin{figure*}[h]
	\centering
    \begin{subfigure}[b]{0.45\textwidth}
		\centering
		\includegraphics[width=\textwidth]{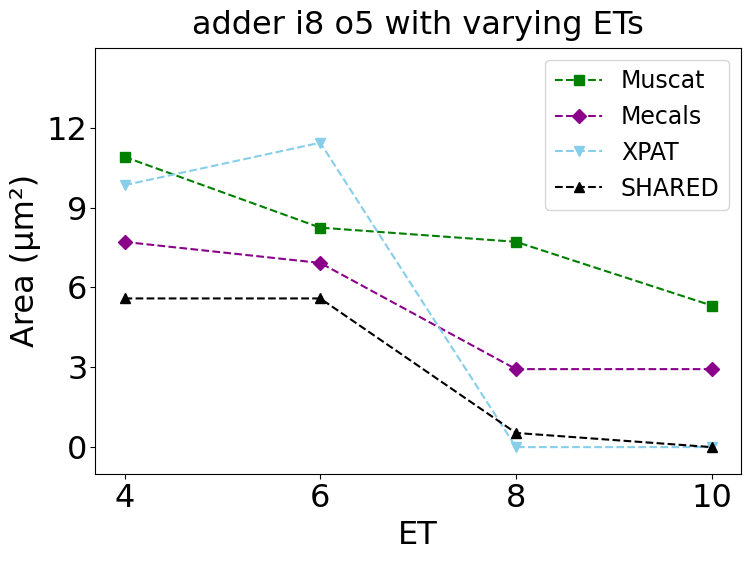}
	\end{subfigure}
    \hfill
	\begin{subfigure}[b]{0.45\textwidth}
		\centering
		\includegraphics[width=\textwidth]{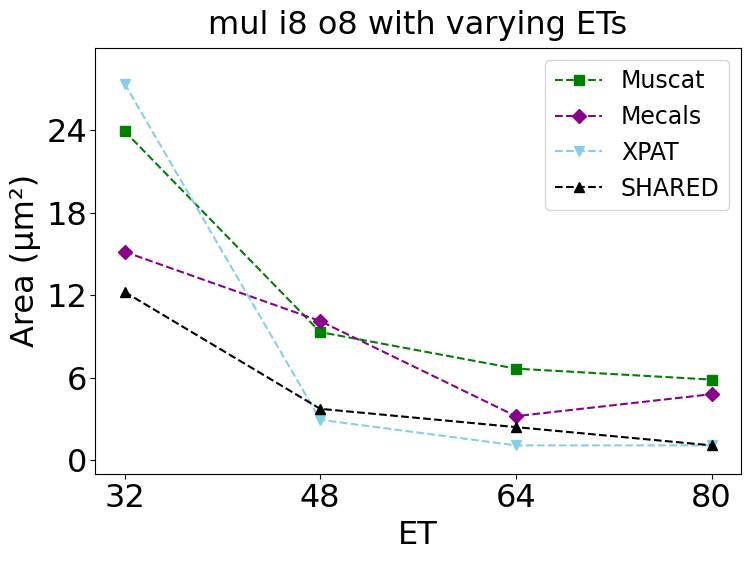}
	\end{subfigure}
	\caption{Area plots for approximate adder and multiplier circuits, showing the area obtained for varying \ac{et} values.}
	\label{fig:area_plots_lines}
\end{figure*}


We implemented our proposed methodology in Python as an open-source tool\footnote{Available at: \url{https://github.com/LP-RG/subxpat}.}. We used \zthree~\cite{Demoura08} v4.11.2 to solve \ac{smt} queries. For synthesis, we employed \yosys~\cite{Yosys} v0.23+21 and the Nangate 45nm cell library. We compared our methodology -- referred to as \shared{} -- against the three best-performing state-of-the-art methods for limiting the \emph{maximum} error of approximate circuits:  \muscat{}~\cite{WitschenMar22}, \mecals{}~\cite{MengMar23}, and XPAT~\cite{Rezaalipour23a}. As benchmarks, we used Verilog specifications of small circuits performing two types of arithmetic operations: addition and multiplication; bitwidths of 2, 3, and 4 were used, indicated in the benchmarks names by i4, i6, and i8. Note that circuits as small as 4-bit multipliers are commonly used for \ac{nn} inference in edge devices~\cite{VenkataramaniJun21}, giving relevance to the (limited) size of circuits considered in our experiments. All experiments had a three hour timeout and were done on a Linux machine with a 3.30 GHz Intel Core i9 and 256 GB of RAM. To assess the impact of our template, we first evaluate the suitability of \ac{pit} and \ac{its} as proxies for synthesised area, and then we study the overall quality of circuits found by SHARED w.r.t. those found by the other methods.

The first set of plots, shown in Fig.~\ref{fig:area_plots_clouds}, reports the results relating proxy values with area, for a fixed \ac{et}, for two adders and two multipliers. The areas of the exact circuits (indicated with a light blue star in the plots) and of 1000 random approximations sound w.r.t. \ac{et} (indicated with a red circle in the plots) are shown to give a baseline with which to judge the compared methods. Note that for the benchmarks with 6 inputs (the bottom row of the figure) the randomly generated approximations are not visible, as they all have area and proxy values which are larger than the scales shown, due to the significant increase in design space size between i4 and i6. The area and proxy values of the circuits produced by the state-of-the-art methods are mostly larger -- sometimes very significantly so -- than those of the circuits produced by SHARED. Note that, for both SHARED and XPAT, several points are reported as opposed to a single one; this is due to the fact that these two methods return several satisfying assignments obtained by the \ac{smt} solver, which we show in order to better illustrate the effect of the proxy. The take-away points from these plots are that (1) \ac{pit} and \ac{its} have a strong correlation with area and (2) SHARED produces circuits that have lower area w.r.t. the state-of-the-art methods.

The second set of experiments, reported in Figure~\ref{fig:area_plots_lines}, shows the areas of the circuits obtained by the four methods, for varying \ac{et} values; in this case only one point, representing the best approximation found, is shown for \xpat{} and for \shared{}. As can be seen, these plots show that SHARED yields the best approximations of the four compared methods for most \ac{et} values.

In all plots, we repo   rt the synthesised area of the circuits obtained, and since power consumption is directly proportional to area, we argue that the circuits we generate are suitable to low-power approximation-resilient applications in general, and to \aclp{nn} on the edge in particular.

%% file: section_conclusion.tex
\section{Conclusion} \label{sec:conclusion}


Approximate computing can help navigating the delicate tradeoff required by \aclp{nn} on the edge, to reach the careful balance between the energy spent for inference and the accuracy of the resulting classification. We postulate that template parameters can act as proxies for metrics of interest in \ac{als}, and propose a novel parametrisable template with the goal of reducing synthesised area in approximate circuits. Our template is based on the sharing of product's outputs on a sum of products structure, and allows the search algorithm to avoid low-quality solutions. We compare our methodology against three state-of-the-art approaches, and our results indicate that the parameters of our template have a strong correlation to synthesised area, and that our methodology outperforms the state of the art in finding smaller circuits given the same \ac{et}.

Avenues for future work include the exploration of templates that (i) are multi-level, (ii) have alternative structures, and (iii) include more complex elements, e.g., \aclp{lut}.